\documentclass[10pt,twocolumn,letterpaper]{article}

\newcommand*\samethanks[1][\value{footnote}]{\footnotemark[#1]}

\usepackage[pagenumbers]{cvpr} % To force page numbers, e.g. for an arXiv version
\usepackage[dvipsnames]{xcolor}

\definecolor{cvprblue}{rgb}{0.21,0.49,0.74}
\usepackage[pagebackref,breaklinks,colorlinks,citecolor=cvprblue]{hyperref}

\usepackage{graphicx}
\usepackage{amsmath}
\usepackage{amssymb}
\usepackage{booktabs}
\usepackage{makecell}
\usepackage{multirow}
\usepackage{bm}
\usepackage{float}
\usepackage[accsupp]{axessibility}

\def\paperID{6365} % *** Enter the Paper ID here
\def\confName{CVPR}
\def\confYear{2024}

\title{Disentangled Pre-training for Human-Object Interaction Detection}

\author{
    Zhuolong Li$^1$\thanks{The first two authors contribute equally.} \quad 
    Xingao Li$^{1}$\samethanks \quad 
    Changxing Ding$^{1,2,3}$\thanks{Corresponding author.} \quad 
    Xiangmin Xu$^{2}$\\
    $^1$School of Electronic and Information Engineering, South China University of Technology\\
    $^2$School of Future Technology, South China University of Technology \quad 
    $^3$Pazhou Lab, Guangzhou\\
{\tt\small \{eezhuolong, eexingao\}@mail.scut.edu.cn,  \{chxding, xmxu\}@scut.edu.cn}
}

\begin{document}
\maketitle
\begin{abstract}
Detecting human-object interaction (HOI) has long been limited by the amount of supervised data available. Recent approaches address this issue by pre-training according to pseudo-labels, which align object regions with HOI triplets parsed from image captions.  However, pseudo-labeling is tricky and noisy, making HOI pre-training a complex process. Therefore, we propose an efficient disentangled pre-training method for HOI detection (DP-HOI) to address this problem. First, DP-HOI utilizes object detection and action recognition datasets to pre-train the detection and interaction decoder layers, respectively. Then, we arrange these decoder layers so that the pre-training architecture is consistent with the downstream HOI detection task. This facilitates efficient knowledge transfer. Specifically, the detection decoder identifies reliable human instances in each action recognition dataset image, generates one corresponding query, and feeds it into the interaction decoder for verb classification. Next, we combine the human instance verb predictions in the same image and impose image-level supervision. The DP-HOI structure can be easily adapted to the HOI detection task, enabling effective model parameter initialization. Therefore, it significantly enhances the performance of existing HOI detection models on a broad range of rare categories. The code and pre-trained weight are available at \href{https://github.com/xingaoli/DP-HOI}{https://github.com/xingaoli/DP-HOI}.
\end{abstract}

\vspace{-4mm}
\section{Introduction}
Human-Object Interaction (HOI) detection involves simultaneous object detection and verb classification for every interactive human-object pair in an image. It is a fundamental scene and action understanding task with various potential applications in robotics~\cite{Mascaro2023HOI4ABOT}, image captioning~\cite{yao2018exploring,pan2020spatio}, image retrieval~\cite{johnson2015image, ding2018trunk}, and visual question answering~\cite{Chen2020Counterfactual}. 
The labeling costs for HOI detection datasets are higher than those for image classification and object detection due to the inclusion of all meaningful $\langle human, verb, object\rangle$ triplets in an image. Therefore, existing HOI detection datasets are usually small, considerably affecting HOI detection performance.

\begin{figure}[t]
  \begin{center}
  \includegraphics[width=1\linewidth]{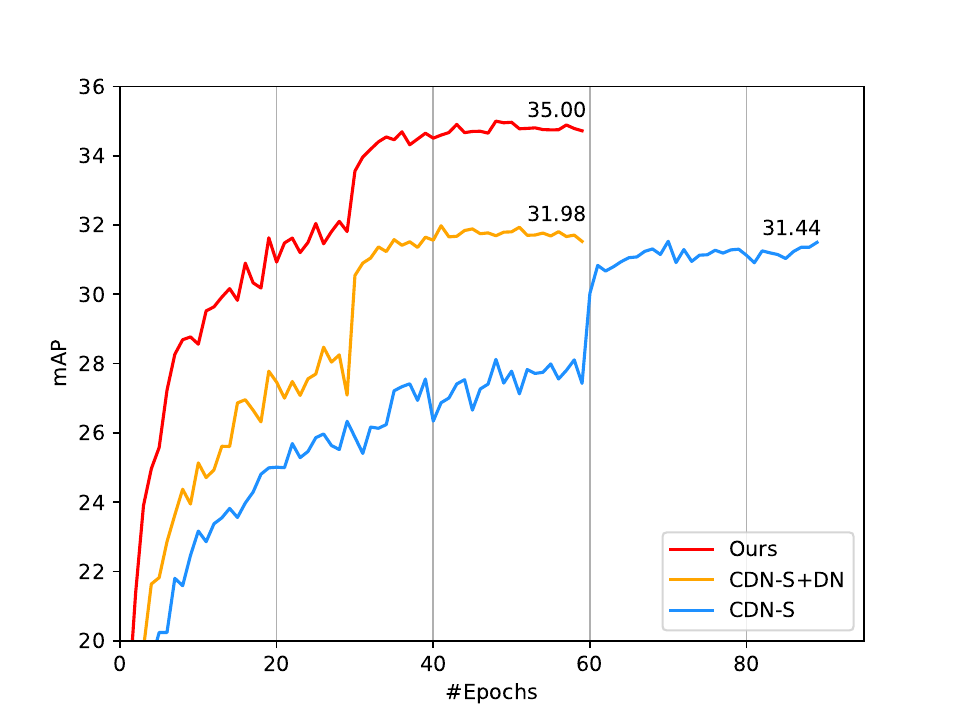}
  \end{center}
  \caption{ 
    CDN-S \cite{zhang2021mining} mAP and convergence curves with the pre-trained DETR weights~\cite{carion2020end} on MS-COCO \cite{lin2014microsoft} and our DP-HOI, respectively. DN denotes the denoising strategy \cite{li2022dn} is adopted to speed up convergence. Experiments are conducted on the HICO-DET dataset~\cite{chao2018learning}.
  }
  \vspace{-4mm}
  \label{fig:intro}
\end{figure}

Current HOI detection models are usually based on detection transformer (DETR)~\cite{carion2020end}. 
Since the DETR training is data hungry, most existing works~\cite{tamura2021qpic,liao2022gen,wang2022chairs,chen2021reformulating,zhou2022human,wu2022mining,liu2022interactiveness} initialize their model according to pre-trained DETR weights on object detection datasets (e.g., MS-COCO~\cite{lin2014microsoft}). 
This strategy is sub-optimal for HOI detection because the pre-trained DETR model does not contain any action knowledge.
As a result, recent studies~\cite{Yuan2023RLIPv2,Zheng2023Open} have adopted large-scale pseudo-labeled scene graph data for HOI model pre-training, which has shown significant potential.

However, the scene graph data pseudo-labeling process is complex and the obtained pseudo-labels are error-prone. 
In this paper, we observed that HOI detection can be decomposed into two sub-tasks: interactive human-object pair detection and interaction classification. These sub-tasks are closely related to the object detection and action recognition tasks, respectively. The labeling of both object detection and action recognition tasks is easier; therefore, they both own large-scale labeled datasets ( e.g., Objects365~\cite{shao2019objects365} and Kinetics-700~\cite{Carreira2019short}). Based on these observations, we propose utilizing these labeled datasets for pre-training HOI detection models. However, these datasets are partially labeled (i.e., only objects or actions are labeled). Thus, they cannot be directly utilized for training according to standard HOI detection architecture. Therefore, a tailored pre-training architecture that is as close as possible to that of the downstream HOI detection task is required for efficient knowledge transfer.

In this study, we propose the disentangled pre-training method for human-object interaction (DP-HOI). DP-HOI conducts object detection and verb classification using two parallel branches. The first branch contains a detection decoder trained with object detection datasets, according to the standard DETR structure and training strategy~\cite{carion2020end,li2022dn}. The second branch is trained using readily available action recognition datasets.

Moreover, we design the verb classification branch to mimic popular HOI detection structures~\cite{Ning_2023_CVPR,liao2022gen,zhang2021mining}. This branch contains a detection and interaction decoder. The detection decoder shares parameters with the object detection branch and identifies all human instances in each training image from the action recognition datasets. Then, we adopt each human instance’s output decoder embedding as a reliable person query (RPQ) for the interaction decoder. Each RPQ is responsible to search for the action cues of the specified person and predict the person's action. 
Since there are only image-level action labels and there might be several RPQs for each image, we introduce a verb-wise prediction fusion (VPF) strategy to merge the RPQ prediction results and impose supervision. In addition, we extend our approach to video and image captioning data, which contain significant action categories vital for pre-training purposes.

Furthermore, we demonstrate DP-HOI’s effectiveness through comprehensive experiments on two popular benchmarks (i.e., HICO-DET and V-COCO), observing that DP-HOI consistently boosts the performance of state-of-the-art HOI detection models. For example, as illustrated in Figure~\ref{fig:intro}, DP-HOI promotes the performance of CDN-S with denoising (DN)~\cite{li2022dn} by 3.02\% mAP.

\section{Related Work}
\subsection{Human-Object Interaction Detection}
%=================================================== 

Existing HOI detection models can be divided into one- and two-stage methods.
Previous two-stage  methods~\cite{Wang_2021_ICCV_Discovering,hou2021affordance,wang2020discovering,hou2020visual,zhong2021polysemy,wang2020contextual,kim2020detecting,Gupta_2019_ICCV,gao2018ican,wang2019deep,zhou2019relation,wan2019pose} employ an off-the-shelf detector to execute object detection before predicting interactions. These methods introduce additional features~\cite{li2020pastanet,zhong2020polysemy,ulutan2020vsgnet,li2020detailed,liu2020consnet,Gupta_2019_ICCV,zhou2019relation,wan2019pose,wang2019deep,gao2018ican,park2023viplo,zhang2023exploring} or external knowledge~\cite{li2019transferable,he2021exploiting,cao2023re} to promote the interaction classification accuracy.
For example, Park et al.~\cite{park2023viplo} propose a pose-conditioned self-loop graph neural network to enhance interaction features, while Cao et al.~\cite{cao2023re} incorporates structured text knowledge to promote HOI detection performance. Due to their multi-stage nature, the two-stage methods generally have a slow inference process. To overcome this problem, one-stage methods~\cite{liao2020ppdm,zhong2021glance,wang2020learning,kim2020uniondet} were proposed and they typically perform object detection and interaction classification in parallel.

Based on DETR’s success~\cite{carion2020end}, recent studies have focused on developing DETR-based HOI detection models, achieving significant performance improvement~\cite{tamura2021qpic,chen2021reformulating,kim2021hotr,zou2021end,zhang2021mining,yuan2022detecting,li2022improving,liao2022gen,zhou2022human, park2022consistency,kim2023relational,Xie2023CVPR,zhong2023disentangled}. This is mainly because the cross-attention operation in transformer decoder layers flexibly extracts image-wide context information for interaction classification. DETR-based methods can be divided into two groups. The first group directly utilizes the conventional DETR structure~\cite{tamura2021qpic,chen2021reformulating,kim2021hotr,zou2021end,zhang2021mining,park2022consistency,zhou2022human}. The second group increases the power of DETR models and can be further divided into two sub-categories: query-enhanced methods~\cite{zhong2022towards,qu2022distillation,dong2022category} and structure-enhanced methods~\cite{zhang2022exploring,iftekhar2022look,wu2022mining,kim2022mstr,zhang2022efficient,tu2022iwin,liao2022gen,Ning_2023_CVPR,tu2023agglomerative}. The query-enhanced methods enhance HOI detection performance with semantically clear queries. In contrast, the structure-enhanced methods aim to develop customized model architectures for HOI detection. 
Some studies~\cite{wan2023weakly,wu2023end,Ning_2023_CVPR,lei2023efficient,liao2022gen} recently proposed improving HOI detection performance by transferring knowledge from visual-linguistic pre-trained models (e.g., CLIP~\cite{radford2021learning}). 

The above methods have achieved impressive performances. However, they still initialize model parameters according to a pre-trained DETR model using object detection datasets. Therefore, their performance in interaction classification may still be sub-optimal.

\subsection{Pre-training Methods for Detection Tasks}
Pre-training and fine-tuning have become popular pipelines for object detection. Due to the DETR architecture’s increase in popularity for object detection, researchers have started studying DETR-specialized pre-training methods. For example, UP-DETR~\cite{dai2021up} utilizes a proxy task that uses a randomly cropped image patch as the query and forces the DETR model to predict the patch location in the image. Moreover, DETReg~\cite{bar2022detreg} uses an unsupervised region proposal generator to produce potential object-bounding boxes. These boxes are used to pre-train the DETR model via the bounding-box regression task.

Since the transformer training is data hungry~\cite{dosovitskiy2020image} and existing HOI detection datasets are usually small~\cite{chao2018learning,gupta2015visual}, DETR-based HOI detection models usually adopt pre-trained DETR weights on object detection datasets. However, this strategy may be unsuitable, since HOI detection includes object detection and interaction classification. To solve this problem, Yuan et al.~\cite{yuan2022rlip} utilized manually labeled scene graph data for HOI detection pre-training. In comparison, subsequent studies~\cite{Zheng2023Open,Yuan2023RLIPv2} proposed various pseudo-labeling approaches that associate image-level HOI labels with object-bounding boxes, significantly expanding the pre-training data scale.

In this paper, we separate the pre-training of the two sub-tasks in HOI detection to bypass the tricky and noisy pseudo-labeling process. In this way, both sub-tasks benefit from clean labels. The experimental results indicate that DP-HOI significantly improves HOI detection performance.

%=====================overview============
\begin{figure*}[!t]
  \begin{center}
  \includegraphics[width=1.0\linewidth]{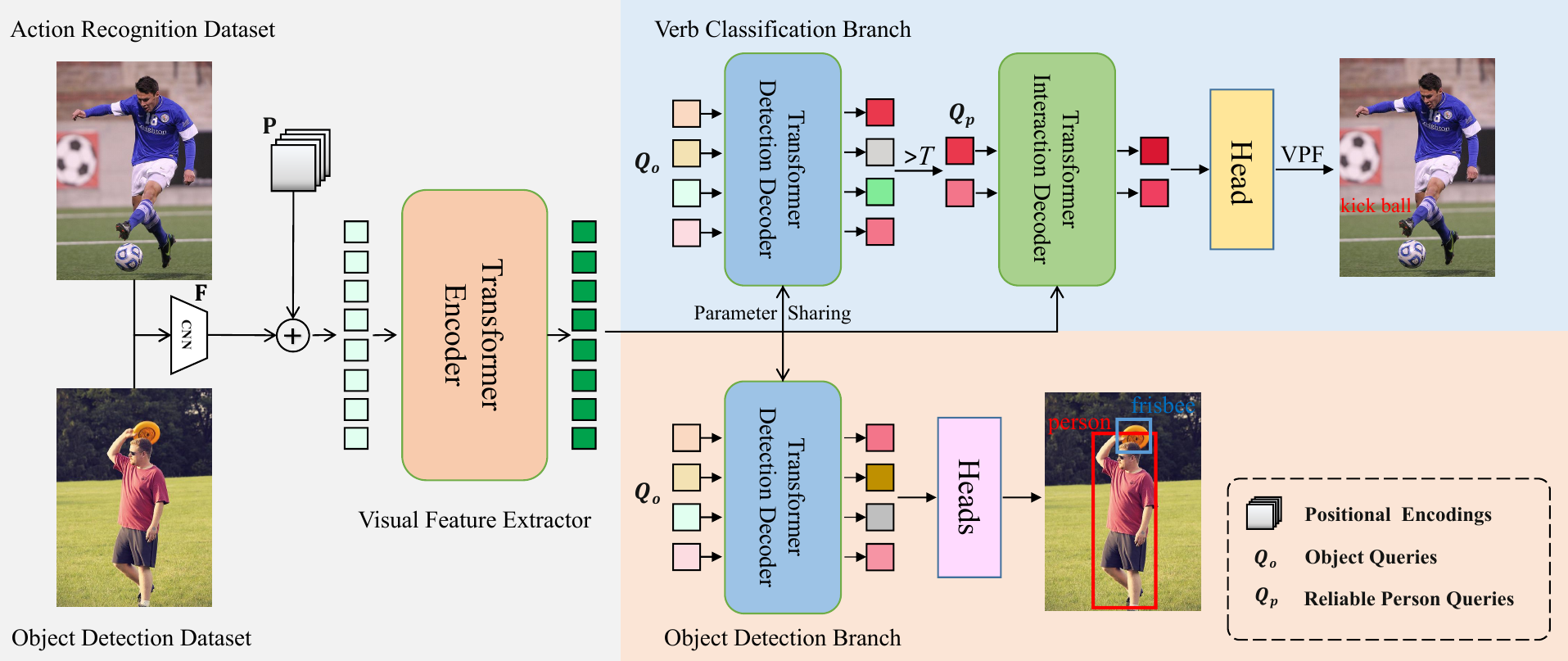}
  \end{center}
  \caption{Our DP-HOI framework overview. It includes a CNN backbone, a transformer encoder, an object detection branch, and a verb classification branch. The two branches are trained in a disentangled manner, with labeled databases for object detection and action recognition, respectively. Each training image from the action recognition dataset first passes the detection decoder, identifies reliable human instances, and generates reliable person queries (RPQs) for the interaction decoder. Then, each RPQ is responsible for searching for relevant action cues for the specified human instance. Since we only have image-level action labels, we impose supervision on the fused RPQs predictions.
  }
  \label{fig:overall}
\end{figure*}
\section{Methods}

In this section, we first briefly describe the research motivation and overall DP-HOI framework. Then, we introduce its detection and verb classification branch structures in Section~\ref{sec:object-detection-branch} and Section~\ref{sec:interaction-classification-branch}, respectively. Moreover, in Section~\ref{sec:data-extension}, we broaden our approach to video-based action and image-caption data. Finally, additional details are provided in Section~\ref{sec:overall-loss}.

\subsection{Overview}\label{sec:overview}
HOI detection can be divided into two sub-tasks: interactive human-object pair detection and interaction classification. These sub-tasks are closely related to the object detection and action recognition tasks. These tasks have large-scale labeled datasets because their annotation costs are cheaper. Moreover, action recognition data can be supplemented with image-caption data. Based on this observation, we proposed using the existing datasets for object detection, verb classification and image captioning to pre-train HOI detection models.

Our proposed DP-HOI framework is illustrated in Figure \ref{fig:overall}. During pre-training, each batch contains a set of images from the object detection ${D_d}$$=$$\{\bm{X}_i^d, \mathbf{y}_i^d\}_{i=1}^{N_d}$ and another set of images from action recognition datasets ${D_a}$$=$$\{\bm{X}_i^a, \mathbf{y}_i^a\}_{i=1}^{N_a}$. $N_d$ and $N_a$ denote the number of images. 
While the annotation $\mathbf{y}_i^d$ contains object bounding boxes and object categories, $\mathbf{y}_i^a$ only contains the verb categories. First, a given image $\bm{X}_i^k$ $(k \in \{d, a\})$ was fed into the CNN backbone in Figure \ref{fig:overall}. Then, the output feature maps were flattened and injected with fixed sine positional encoding. Finally, the feature maps were enhanced by the self-attention operations in the transformer encoder.

Since $\mathbf{y}_i^d$ and $\mathbf{y}_i^a$ contain object and action labels only, we utilized the enhanced $\bm{X}_i^d$ and $\bm{X}_i^a$ features in a disentangled manner. In summary, there are two branches after the transformer encoder, i.e., one detection branch and one verb classification branch. The enhanced $\bm{X}_i^d$ and $\bm{X}_i^a$ features pass through the detection and verb classification branches, respectively.

\subsection{The Object Detection Branch}\label{sec:object-detection-branch}
This branch contains a detection decoder. We denote the features enhanced by the transformer encoder as $\textbf{V}_e$, the learnable object queries as $\mathbf Q_o=\left\{Q_0, Q_1, ..., Q_{N-1}\right\}$, and the initial decoder embeddings as $\mathbf{o_{0}}$. The output decoder embeddings $\mathbf{o_{d}}$ can be represented as follows:

\vspace{-2mm}
\begin{equation}
\centering
    \mathbf{o_{d}}=D_{d}
    (\mathbf{Q_o},\mathbf{o_{0}, \mathbf{V}_e}),
\end{equation}
where $D_{d}$($\cdot$) represents the detection decoder. Finally, $\mathbf{o_{d}}$ was employed to predict object bounding boxes and categories using feed-forward networks (FFNs):

\vspace{-2mm}
\begin{equation}
\centering
\vspace{-2mm}
 {\mathbf{\hat{y}}_{box}} = 
 {f_{h}}(\mathbf{o_{d}}),
\vspace{-2mm}
\end{equation}

\begin{equation}
\centering
 {\mathbf{\hat{y}}_{o}} = 
 {f_{o}}(\mathbf{o_{d}}),
\end{equation}
where $f_{h}$ and $f_{o}$ denote two FFNs. Finally, we imposed supervision on $\mathbf{\hat{y}}_{box}$ and $\mathbf{\hat{y}}_{o}$ via bipartite matching~\cite{carion2020end}.

\subsection{The Verb Classification Branch}\label{sec:interaction-classification-branch}

We designed the DP-HOI verb classification branch to maximize the pre-training efficacy according to the structure of recently popular HOI detection models (e.g., CDN~\cite{zhang2021mining}). In the experimentation section, we demonstrate that DP-HOI significantly improves the performance of other HOI detection models.

The verb classification branch contains two sequential decoders (i.e., a detection decoder and interaction decoder). The same as~\cite{zhang2021mining,liao2022gen}, the first decoder’s output embeddings were utilized as the queries for the second one.
The detection decoder shares parameters with that in the object detection branch. The two decoders are utilized for object detection and verb classification, respectively.

A slight difference exists between our detection decoder and that in the CDN~\cite{zhang2021mining} model.
Specifically, the CDN model’s detection decoder detects interactive human–object pairs, enabling its interaction decoder to recognize the verb categories of a specific human–object pair. In comparison, our adopted pre-training datasets do not contain any interactive human–object pair annotations. Hence, we proposed reducing the action recognition within a human–object pair to identifying all the actions performed by a human instance. Thus, we selected reliable human instances according to the detection decoder’s predictions.

\noindent\textbf{Reliable Person Queries}.
In this study, a human instance was regarded as reliable if the detection decoder’s human category prediction score was above the threshold $T$. RPQs are the decoder embeddings in $\mathbf{o_{d}}$ that predict these reliable instances for the interaction decoder. The collection of the RPQs for one image is denoted as $\mathbf{Q_p}$. Each RPQ searches for action-relevant cues using cross-attention on the specific person within the interaction decoder:

\vspace{-1mm}
\begin{equation}
    \mathbf{o_{a}}=D_{a}
    (\mathbf{Q_p},\mathbf{o_{0}, \mathbf{V}_e}),
\end{equation}
where $D_{a}$($\cdot$) represents the interaction decoder and $\mathbf{o_{a}}$ denotes the output decoder embeddings from the interaction decoder. Finally, $\mathbf{o_{a}}$ is utilized for verb classification:

\vspace{-2mm}
\begin{equation}
 {\mathbf{\hat{y}}_{a}} = 
 {f_{a}}(\mathbf{o_{a}}),
\end{equation}
where $f_{a}$ denotes an FFN and $\mathbf{\hat{y}}_a \in \mathbb{R}^{N_p \times C_a}$. $N_p$ and $C_a$ represent the number of RPQs and the number of verb classes, respectively. 

\noindent\textbf{Verb-wise Prediction Fusion}. Although action recognition datasets generally do not provide instance-level action annotations, a single image may contain multiple human instances. Hence, we propose Verb-wise Prediction Fusion (VPF) to fuse the prediction results in $\mathbf{\hat{y}}_a$ by conducting \textit{max-pooling} along its column dimension. In the experimentation section, we demonstrated that VPF outperforms other fusion strategies and effectively suppresses noisy RPQ predictions.

\subsection{Extension to Video and Caption Data}\label{sec:data-extension}
\noindent\textbf{Video Data}.  
Existing action recognition datasets are usually video-based. 
To utilize these video-based datasets, we randomly sampled $N_f$ frames from each video and fed them into our model to obtain RPQ prediction results, denoted by $\mathbf\{{\hat{y}}_a\}_{N_f}$. Then, we utilized the VPF method to fuse these prediction results. 
Finally, we adopted focal loss~\cite{lin2017focal} to supervise the fused results according to the video label.

\noindent\textbf{Image-Caption Data}. Since action recognition datasets are labeled according to fixed action categories, the action semantics they contain are insufficient. As a result, contrastive learning was utilized, enabling the use of image-caption data with robust action semantic information for pre-training. First, we used a rule-based language parser~\cite{qi2018learning} to obtain HOI triplets $\langle human, verb, object\rangle$ for a given image-caption pair. Then, we fed each HOI-triplet prompt (i.e., a photo of \{human\} \{verb\} \{object\}) into the CLIP text encoder to obtain its embedding. 

Selecting negative samples during contrastive learning significantly impacts model performance. In this paper, we clustered all the HOI-triplet text embedding categories into 100 offline clusters. Then we sampled 10 HOI categories from each cluster as negative samples for each corresponding RPQ embedding. Next, we calculated the cosine similarity between the RPQs’ decoder and text embeddings, selecting the RPQ with the highest similarity score.

Finally, we compute the InfoNCE loss \cite{oord2018representation} separately in two directions to obtain the image alignment loss $\mathcal{L}_{i2t}$ and the text alignment loss $\mathcal{L}_{t2i}$. The average of these two loss functions is used as the final loss $\mathcal{L}_{s}$:

\vspace{-2.5mm}
\begin{equation}
 \mathcal{L}_{s} = \frac{1}{2}(\mathcal{L}_{i2t} + \mathcal{L}_{t2i}).
\label{eq:loss_ct}
\end{equation}

\subsection{Overall Loss Function}\label{sec:overall-loss}
We adopted similar loss functions as existing object detection~\cite{carion2020end} and verb classification~\cite{tamura2021qpic} studies. The overall DP-HOI loss function is represented as follows:

\vspace{-2.5mm}
\begin{equation}
 \mathcal{L} = \mathcal{L}_{d} + \lambda_{v} \mathcal{L}_{v},
\label{eq:loss_all}
\end{equation}

\vspace{-2.5mm}
\begin{equation}
 \mathcal{L}_{d} = \lambda_{b} \mathcal{L}_{b} + \lambda_{g} \mathcal{L}_{g} + \lambda_{c} \mathcal{L}_{c},
\label{eq:loss_detection}
\end{equation}

\vspace{-2.5mm}
\begin{equation}
 \mathcal{L}_{v} = \lambda_{a} \mathcal{L}_{a} + \lambda_{s} \mathcal{L}_{s},
\label{eq:loss_verb}
\end{equation}
where $\mathcal{L}_{d}$ and $\mathcal{L}_{v}$ denote the object detection and verb classification branches’ loss functions, respectively.
$\mathcal{L}_{b}$, 
$\mathcal{L}_{g}$,
$\mathcal{L}_{c}$,
$\mathcal{L}_{a}$
and
$\mathcal{L}_{s}$ represent the losses, including L1 and GIOU~\cite{rezatofighi2019generalized} for bounding box regression, cross-entropy for object classification,
focal~\cite{lin2017focal} and InfoNCE~ \cite{oord2018representation} for verb prediction, respectively. In addition,
$\lambda_{v}$ is a weight that balances the 
two branches’ losses.
$\lambda_{b}$, $\lambda_{g}$, $\lambda_{c}$, $\lambda_{a}$ and $\lambda_{s}$ are set as 5, 2, 1, 1 and 1, respectively.

Moreover, we utilized multiple action recognition and image-caption datasets for pre-training. $C_a$ denotes the total verb category number of all the action recognition datasets. Since semantically overlapping verb categories may exist between different datasets, we only activate the binary classifiers for the verb categories owned by the database that each training sample belongs to. More pre-training details are provided in the supplementary material.

\section{Experiments}
\subsection{The Pre-training Datasets}

As illustrated in Table~\ref{table:datasets}, we adopted the MS-COCO~\cite{lin2014microsoft} and Objects365~\cite{shao2019objects365} datasets for the object detection branch. Then, we employed the action recognition and image-caption datasets in the verb classification branch. First, the action recognition datasets included
Haa500~\cite{chung2021Haa500} and Kinetics-700~\cite{Carreira2019short}.
Haa500 and Kinetics-700 are video-based datasets; therefore, we sampled frames at regular intervals during data processing. Considering the lower quality of video frames in Kinetics-700 compared to Haa500, we treated each sampled frame from Haa500 as an individual supervision sample. Then, we applied video-level supervision to the frame sequences sampled from Kinetics-700. Second, the image-caption datasets included
Flickr30k~\cite{young2014image} and VG~\cite{krishna2017visual}. Aside from captions, Flickr30k and VG include additional annotation information; however, we only used the caption annotations for our pre-training. We also filtered the images from which caption HOI triplets could not be extracted. The datasets and data processing methods are detailed in the supplementary material.

\begin{table}[tp]
\caption{Statistics of the adopted object detection, action recognition and image-caption datasets used for pre-training.}
\vspace{-1em}
\label{table:datasets}
\centering
\resizebox{0.45\textwidth}{!}{%
\begin{tabular}{c|c|cc}
\toprule
Types & Datasets & \#Samples & \#Classes   \\ \midrule \midrule 
\multirow{2}{*}{Object Detection} & COCO & 117266 & 80  \\ 
 & Objects365 & 117266 & 365  \\ \midrule
 \multirow{2}{*}{Action Recognition} & Haa500 & 52644 & 500  \\ 
 & Kinetics-700 & 117266 & 700 \\ 
 \midrule
 \multirow{2}{*}{Image-Caption}
 & Flickr30k & 25977 & - \\  
 & VG & 54280 & - \\ 
 \bottomrule
\end{tabular}}
\vspace{-3mm}
\end{table}

\subsection{The HOI Detection Datasets}
\noindent\textbf{HICO-DET.} HICO-DET \cite{chao2018learning} is a popular dataset for HOI detection.
It consists of 47,776 images with more than 150,000 human-object pairs, 38,118 of which are used for training and 9,658 for testing.
This dataset contains the same 80 object classes as MS-COCO~\cite{lin2014microsoft} and 117 interaction classes. The combination of object and interaction classes forms 600 HOI categories. Also, there are 138 HOI categories with less than 10 training samples, which are denoted as “rare” categories. We conducted experiments using the default(DT) mode and three zero-shot settings (i.e., UV, RF-UC, and NF-UC). UV and UC represent unseen verb and composition settings, respectively. RF means rare first, and NF is non-rare first.

\noindent\textbf{V-COCO.} V-COCO \cite{gupta2015visual} is a relatively small dataset, built on the MS-COCO database~\cite{lin2014microsoft}. It contains 10,346 images (i.e., 5,400 for training and 4,946 for testing), covering the same 80 object categories as MS-COCO~\cite{lin2014microsoft} and 26 interaction categories. We use the mean average precision of Scenario 1 ($\rm AP_{role}$)~\cite{gupta2015visual} for evaluation.

\subsection{Implementation Details}

We adopted ResNet-50 as our backbone model. We utilized the AdamW~\cite{loshchilov2017decoupled} optimizer to conduct experiments with a batch size of 64 on 8 A800 GPUs for DP-HOI. In each batch, the number of samples from the object detection and action recognition datasets is equal.
The initial learning rate was set to 1e-4 and multiplied by 0.1 after 180 epochs. The pre-training stage lasts for 200 epochs according to the MS-COCO dataset. Regarding the Kinetics-700 samples, we resized the input video frames from their original size to 256×256 pixels. Meanwhile, the other datasets' input samples were resized to a minimum and maximum of 800 pixels and 1,333 pixels on the short and long sides, respectively.
$\lambda_{v}$, $T$, and $N$ were set to 1, 0.9, and 100, respectively. 
Furthermore, we adopted the DN~\cite{li2022dn} strategy to accelerate the pre-training stage, and the number of detection and interaction decoder layers was set to 3. Please refer to the supplementary material for more implementation details.

%%%%%%HICO%%%%%%%%%%%%%%%%%%%%%%%%%
\begin{table}[t]
\begin{center}
\large
\caption{Performance comparisons for HICO-DET. \textsuperscript{\dag} means DN was adopted in the fine-tuning stage. * denotes a data augmentation strategy~\cite{qu2022distillation} was employed.}
\resizebox{0.45\textwidth}{!}{
\begin{tabular}{c|c|ccc}
\toprule
\multirow{2}{*}{Methods}  & \multirow{2}{*}{Backbone}  & \multicolumn{3}{c}{DT Mode} \\
 && Full & Rare & Non-Rare \\
\midrule
\midrule 
InteractNet \cite{gkioxari2018detecting} & ResNet-50-FPN   &9.94  & 7.16  & 10.77  \\
GPNN\cite{qi2018learning} & Res-DCN-152   &13.11  & 9.34  & 14.23  \\
iCAN \cite{gao2018ican} & ResNet-50   &14.84  & 10.45  & 16.15   \\
No-Frills \cite{Gupta_2019_ICCV} &ResNet-152 	&17.18	&12.17	&18.68 \\ 
UnionDet \cite{kim2020uniondet}& ResNet-50-FPN  &17.58  & 11.72  & 19.33\\
DRG \cite{gao2020drg}	&ResNet-50-FPN			&19.26	&17.74	&19.71 \\
PD-Net \cite{zhong2020polysemy}  & ResNet-152  &20.81 &15.90 &22.28 \\
PPDM \cite{liao2020ppdm}& Hourglass-104   & 21.73 &13.78 &24.10 \\
GGNet \cite{zhong2021glance}    & Hourglass-104   &23.47 &16.48 &25.60 \\
HOTR \cite{kim2021hotr}  & ResNet-50 &23.46 &16.21 &25.62\\
HOI-Trans \cite{zou2021end}   & ResNet-50  & 23.46 & 16.91 & 25.41 \\
 AS-Net \cite{kim2020uniondet}   & ResNet-50  &28.87  & 24.25  & 30.25 \\
QPIC  \cite{tamura2021qpic}  & ResNet-50   &29.07 &21.85 &31.23 \\
CDN-S \cite{zhang2021mining}     & ResNet-50  &31.44 & 27.39 & 32.64 \\
HQM(CDN-S) \cite{zhong2022towards} & ResNet-50  & 32.47 & 28.15 & 33.76\\
DOQ(CDN-S) \cite{qu2022distillation} &ResNet-50  &33.28 &29.19 &34.50 \\
GEN-VLKT$_s$ \cite{liao2022gen} & ResNet-50  &33.75&29.25&35.10 \\
\midrule
\multicolumn{5}{l}{\textit{with our pre-trained model weights}} \\
\midrule
    UPT \cite{zhang2022efficient}  & ResNet-50  & 31.66 & 25.94 & 33.36 \\
    UPT + Ours  & ResNet-50   & \textbf{33.36}  & \textbf{28.74}  & \textbf{34.75}   \\
\midrule
    PViC \cite{zhang2023exploring} & ResNet-50   & 34.69  & 32.14  & 35.45  	 \\
    PViC + Ours  & ResNet-50   & \textbf{35.77}  & \textbf{32.26}  & \textbf{36.81}   \\
\midrule
CDN-S\textsuperscript{\dag} \cite{zhang2021mining}    & ResNet-50  &31.98 & 28.61 & 32.99  \\
CDN-S\textsuperscript{\dag}+Ours &ResNet-50 &35.00 &32.38  &\textbf{35.78}	\\
CDN-S\textsuperscript{\dag}+CCS\textsuperscript{$*$}+Ours &ResNet-50  &\textbf{35.38} &\textbf{34.61}	&35.61\\
\midrule
HOICLIP \cite{{Ning_2023_CVPR}} & ResNet-50  &34.69&31.12&35.74 \\
HOICLIP+Ours &ResNet-50  &\textbf{36.56} &\textbf{34.36}	&\textbf{37.22}	\\
\midrule
\multicolumn{5}{l}{\textit{comparison with pre-training methods}} \\
\midrule
OpenCat (754k)~\cite{Zheng2023Open} &ResNet-101  &32.68 &28.42&33.75 \\
RLIP (225k)~\cite{yuan2022rlip} &ResNet-50  &32.84 &26.85&34.63 \\
RLIPv2 (1,967k)~\cite{Yuan2023RLIPv2} &ResNet-50  &35.38 &29.61&37.10 \\
CDN-S\textsuperscript{\dag}+CCS\textsuperscript{$*$}+Ours (484k) &ResNet-50  &35.38 &\textbf{34.61}	&35.61	\\
HOICLIP+Ours (484k) &ResNet-50  &\textbf{36.56} &34.36	&\textbf{37.22}	\\
\bottomrule
\end{tabular}}
\label{tab:hico}
% \vspace{-7mm}
\end{center}
\end{table}
%%%%%%%%%%%%%%%%%%%%%%%%%%%%%%%%%%%%%%%%%%%%

\subsection{Comparisons with State-of-the-Art Methods}
\noindent{\bf{Comparisons on HICO-DET under default setting}}. 
We directly applied the pre-trained DETR weights from DP-HOI to existing popular methods.

As shown in Table \ref{tab:hico}, DP-HOI significantly enhances the performance of one- and two-stage HOI detection methods.
When the pre-trained DP-HOI weights are applied to two-stage methods, UPT~ \cite{zhang2022efficient} and PViC~\cite{zhang2023exploring}, consistent performance gains of 1.70\% and 1.08\% mAP were observed in DT mode for the full categories, respectively. Furthermore, applying pre-trained DP-HOI weights to one-stage methods, CDN-S\textsuperscript{\dag}~ \cite{zhang2021mining} and HOICLIP~\cite{{Ning_2023_CVPR}}, yielded consistent performance improvements of 3.02\% and 1.87\% mAP, respectively.
Notably, we observed remarkable performance improvements on the rare HOI categories.
For example, compared with DOQ \cite{qu2022distillation}, which also adopts CCS as the data augmentation method, the performance of CDN-S\textsuperscript{\dag} on the rare categories was promoted by 5.42\% mAP, reaching 34.61\% mAP. 
These improvements demonstrated DP-HOI's efficiency and universality.

Moreover, compared with other HOI pre-training methods, DP-HOI outperforms OpenCat~\cite{Zheng2023Open} and RLIP~\cite{yuan2022rlip}.
Its performance was similar to RLIPv2~\cite{Yuan2023RLIPv2} with less pre-training data.
RLIPv2 introduces a complex scheme to obtain pseudo-labeled scene graph data from object detection datasets and adopts 1,967k images for pre-training.
In contrast, DP-HOI adopts a concise pre-training strategy that effectively leverages action semantic information from action recognition and image-caption datasets. It also yields similar results with less data for the simple baseline CDN-S\textsuperscript{\dag}+CCS\textsuperscript{$*$}.
With a stronger baseline HOICLIP \cite{Ning_2023_CVPR}, DP-HOI outperforms RLIPv2, especially on the rare HOI categories.

%%%%%%%%%%%zero-shot%%%%%%%%%%% 
\begin{table}[t]
    \caption{Performance comparisons with state-of-the-art methods for zero-shot HOI detection on HICO-DET. UV and UC indicate unseen verb and composition settings, respectively. RF is short for rare first. NF is non-rare first.
    }
    \label{table:zero-shot}
    \centering
    \large
     \resizebox{0.38\textwidth}{!}{
      \begin{tabular}{c|c|ccc}
        \toprule
        Methods & Type & Unseen & Seen & Full  \\
        \midrule
        \midrule
        GEN-VLKT$_s$ \cite{liao2022gen} & UV & 20.96 & 30.23 & 28.74  \\
        EoID \cite{wu2023end} & UV & 22.71 & 30.73 & 29.61  \\
        OpenCat \cite{Zheng2023Open} & UV & 19.48 & 29.02 & 27.43 \\
        HOICLIP \cite{Ning_2023_CVPR} & UV & 24.30 & 32.19 & 31.09  \\
        HOICLIP+Ours & UV & \textbf{26.30} & \textbf{34.49} & \textbf{33.34} \\
        \midrule
        % VCL \cite{hou2020visual} & RF-UC & 10.06 & 24.28 & 21.43 \\
        % ATL \cite{hou2021affordance} & RF-UC & 9.18 & 24.67 & 21.57 \\
        GEN-VLKT$_s$ \cite{liao2022gen} & RF-UC & 21.36 & 32.91 & 30.56  \\
        EoID \cite{wu2023end} & RF-UC & 22.04 & 31.39 & 29.52  \\
        OpenCat \cite{Zheng2023Open} & RF-UC & 21.46 & 33.86 & 31.38 \\
        RLIPv2 \cite{Yuan2023RLIPv2} & RF-UC & 21.45 & 35.85 & 32.97 \\
        HOICLIP \cite{Ning_2023_CVPR} & RF-UC & 25.53 & 34.85 & 32.99  \\
        HOICLIP+Ours & RF-UC & \textbf{30.49} & \textbf{36.17} & \textbf{35.03} \\
        \midrule
        % VCL \cite{hou2020visual} & NF-UC & 16.22 & 18.52 & 18.06 \\
        % ATL \cite{hou2021affordance} & NF-UC & 18.25 & 18.78 & 18.67 \\
        GEN-VLKT$_s$ \cite{liao2022gen} & NF-UC & 25.05 & 23.38 & 23.71  \\
        EoID \cite{wu2023end} & NF-UC & 26.77 & 26.66 & 26.69  \\
        OpenCat \cite{Zheng2023Open} & NF-UC & 23.25 & 28.04 & 27.08 \\
        RLIPv2 \cite{Yuan2023RLIPv2} & NF-UC & 22.81 & 29.52 & 28.18 \\
        HOICLIP \cite{Ning_2023_CVPR} & NF-UC & 26.39 & 28.10 & 27.75  \\
        HOICLIP+Ours & NF-UC & \textbf{28.87} & \textbf{29.98} & \textbf{29.76} \\
        \bottomrule 
    \end{tabular}
     }
     \vspace{-4mm}
\end{table}
%%%%%%%%%%%%%%%%%%

\noindent{\bf{Comparisons on HICO-DET under zero-shot settings}}. 
To further demonstrate DP-HOI's effectiveness, we conducted experiments using various zero-shot settings, including UV, RF-UC, and NF-UC.

As illustrated in Table \ref{table:zero-shot}, DP-HOI achieved competitive performance across all three zero-shot settings. It outperforms state-of-the-art methods under the UV, RF-UC and NF-UC settings, reaching 26.30\%, 30.49\%, and 28.87\% mAP in the unseen categories, respectively.
In contrast to HOICLIP~\cite{Ning_2023_CVPR}, we observed consistent performance gains of 2.00\%, 4.96\% and 2.48\% mAP for unseen categories under the UV, RF-UC and NF-UC settings.

%%%%%%vcoco%%%%%%%%%%%%%%%%%%%%%%%%%
\begin{table}[htbp]
\begin{center}
\large
\caption{Performance comparisons for V-COCO. GEN$_s$ indicates that CLIP distillation was removed from GEN-VLKT$_s$.}
\label{tab:v-coco}
\resizebox{0.35\textwidth}{!}{
\begin{tabular}{c|c|c}
\midrule
Methods    & Backbone    & $AP_{role}$ \\
\midrule
\midrule
InteractNet \cite{gkioxari2018detecting} &ResNet-50-FPN  &40.0 \\
DRG \cite{gao2020drg}  &ResNet-50-FPN  &51.0 \\
PD-Net \cite{zhong2020polysemy} &ResNet-152  &52.6\\
IDN \cite{li2020hoi} &ResNet-50-FPN  &53.3 \\
GGNet \cite{zhong2021glance}    &Hourglass-104  &54.7 \\
HOTR \cite{kim2021hotr} &ResNet-50  &55.2 \\
HOI-Trans \cite{zou2021end} &ResNet-101  &52.9 \\
AS-Net \cite{chen2021reformulating} &ResNet-50&53.9\\
IF \cite{liu2022interactiveness} &ResNet-50  &63.0\\
PartMap \cite{wu2022mining} &ResNet-50  &63.0\\
DOQ(QPIC) \cite{qu2022distillation}  & ResNet-50 &63.5 \\
HQM(QPIC) \cite{zhong2022towards}  & ResNet-50 &63.6 \\
\midrule
\multicolumn{3}{l}{\textit{with our pre-trained model weights}} \\
\midrule
QPIC \cite{tamura2021qpic} & ResNet-50 &58.8\\
QPIC+Ours &ResNet-50  &\textbf{63.2}\\
\midrule
CDN-S \cite{zhang2021mining} & ResNet-50 &61.7\\
CDN-S+Ours &ResNet-50  &\textbf{64.8}\\
\midrule
GEN$_s$+VLKT \cite{liao2022gen} & ResNet-50  &62.4 \\
GEN$_s$+Ours &ResNet-50 &\textbf{66.6}\\
\midrule
\multicolumn{3}{l}{\textit{comparisons with pre-training methods}} \\
\midrule
OpenCat (754k) \cite{Zheng2023Open} &ResNet-50  &61.9\\
RLIP (225k)~\cite{yuan2022rlip} &ResNet-101  &61.9\\
RLIPv2 (1,967k)~\cite{Yuan2023RLIPv2} &ResNet-50  &65.9 \\
GEN$_s$+Ours (484k) &ResNet-50  &\textbf{66.6}\\
\bottomrule
\end{tabular}}
\vspace{-3mm}
\end{center}
\end{table}

\noindent{\bf{Comparisons on V-COCO}}. Table~\ref{tab:v-coco} displays the V-COCO comparisons. We observed that DP-HOI consistently enhances the model's performance on the V-COCO dataset, reaching 63.2\%, 64.8\%, and 66.6\% mAP in the $AP_{role}$ for QPIC~\cite{tamura2021qpic}, CDN-S~\cite{zhang2021mining}, and GEN$_s$~\cite{liao2022gen}, respectively.
GEN$_s$+Ours method outperforms other HOI pre-training approaches with 484k samples.
These results demonstrate that DP-HOI provides superior pre-trained weights for HOI detection models. 

\subsection{The Experiment Using Different Datasets}
We conducted experiments using various pre-training data combinations to explore their impact on the datasets, as illustrated in Table~\ref{table:more_datasets_results}. COCO indicates pre-training with only the MS-COCO dataset, which is regarded as the baseline. ALL signifies pre-training with the 484k data shown in Table~\ref{table:datasets}.

Initially, we extended pre-training data from the Haa500, Kinetics-700, and Flickr30k datasets independently, reaching 1.34\%, 1.36\% and 1.33\% improvement on full categories. 
These results demonstrate that using various action dataset types (e.g., action image, action video and image-caption dataset) could be beneficial for DP-HOI.
Furthermore, we observed consistent performance improvements when integrating diverse action datasets.
With the 484k pre-training data, the DP-HOI outperforms the baseline by 3.02\%, 3.77\%, and 2.79\% mAP on the full, rare, and non-rare HOI categories, respectively.
These experimental results demonstrate DP-HOI's scalability and effectiveness. Moreover, we encountered an intriguing observation: integrating image-caption data led to a remarkable improvement on the non-rare categories. This could be attributed to the diverse range of action classes in the image-caption dataset.

% %Data scale 
\begin{table}[htbp]
    \caption{Performance comparisons using different pre-training datasets.
    }
    \label{table:more_datasets_results}
    \centering
    \small
     \resizebox{0.35\textwidth}{!}{
      \begin{tabular}{c|ccc}
        \toprule
        Datasets & Full & Rare & Non-Rare  \\
        \midrule
        \midrule
        COCO & 31.98 & 28.61 & 32.99  \\
        \midrule
        +Haa500 & 33.32 & 30.18 & 34.26\\
        +Kinetics-700 & 33.34 & 29.89 & 34.37\\
        +Flickr30k & 33.31&30.76 &34.08 \\
        +Haa500, Flickr30k & 34.06&30.89 &35.01 \\
        +Flickr30k, VG & 33.77&31.54 &34.43 \\
        \midrule
        ALL &\textbf{35.00} &\textbf{32.38} &\textbf{35.78}\\    
        \bottomrule 
    \end{tabular}
     }
\end{table}

\subsection{The Ablation Study}
 We conducted pre-training using the COCO and Kinetics-700 datasets in the ablation study. Then, we fine-tuned the HICO-DET database with the CDN-S model~\cite{zhang2021mining} adopted through DN~\cite{li2022dn}. The detailed experimental results are summarized in Table~\ref{tab:ablation_study_1}. COCO indicates pre-training with only the MS-COCO dataset, which is regarded as the baseline. K700 represents the 117k Kinetics-700 data in Table~\ref{table:datasets}.

%%%%%%%%%%%%%%ablation-study%%%%%%%%%%%%%%%%%%%%%
\begin{table}[t]
\large
	\centering
\caption{Ablation study on each key DP-HOI component.}
	\resizebox{0.45\textwidth}{!}{
		\begin{tabular}{c cccc ccc}
		\toprule
		&\multicolumn{4}{c}{Components}  &\multicolumn{3}{c}{mAP}  \\
		Methods  &COCO &K700 &VPF &RPQ  & Full &Rare &Non-Rare\\
		\midrule
		\multirow{1}{*}{Baseline}
        &\checkmark&-&-&-&31.98&28.61 &32.99 \\
        \midrule
        \multirow{3}{*}{Incremental}
         &\checkmark&\checkmark&-&-&32.69&28.64&33.90 \\
         &\checkmark&\checkmark&\checkmark&-&32.93&28.87&34.14 \\
	&\checkmark&\checkmark&-&\checkmark&33.07&28.96&34.30\\
        \midrule
        \multirow{1}{*}{Ours}
&\checkmark&\checkmark&\checkmark&\checkmark&\textbf{33.34}&\textbf{29.89}&\textbf{34.37}\\
		\bottomrule
	\end{tabular}}
        \label{tab:ablation_study_1}
        % \vspace{-2mm}
\end{table}

\noindent\textbf{The Effectiveness of DP-HOI.}
As illustrated in Table \ref{tab:ablation_study_1}, DP-HOI significantly outperforms the baseline by 1.36\%, 1.28\%, and 1.38\% mAP in DT mode for the full, rare, and non-rare HOI categories, respectively.

\noindent\textbf{The Effectiveness of VPF.}
`max-pooling' was proposed to fuse the RPQ prediction results. 
As shown in Table~\ref{tab:ablation_study_1}, when VPF was removed, the HICO-DET full category performance declined by 0.27\% mAP. 
Since the people in the video frames may not have performed any action, directly imposing supervision on each RPQ without fusion is unreasonable.

\noindent\textbf{The Effectiveness of RPQ.}
RPQ was employed to identify human instances from the detection decoder and generate person-specific queries for the subsequent interaction decoder. This strategy enabled the model to focus on the action cues for each specific person. When RPQ was removed and all detection decoder embeddings were fed into the interaction decoder, the HOI detection performance decreased by 0.41\% mAP. The above experimental results verify that query selection for the interaction decoder is essential in DP-HOI.

\subsection{Comparisons With DP-HOI Variants}\label{sec:Variants}
\noindent\textbf{The Comparisons With VPF Variants.} We compared the VPF's performance with two of its variants. The experimental results are displayed in Table~\ref{table:VPF_ablation_results}. The first variant is denoted as ``w/o fusion'', indicating that we imposed supervision on each confident RPQ. As a result, it displayed a performance lower than VPF by 0.27\%, 0.93\%, and 0.07\% mAP in DT mode for the full, rare, and non-rare HOI categories, respectively. We estimate that conducting max-pooling on all RPQ predictions enables VPF to suppress noisy predictions by non-confident RPQs.

\begin{table}[t]
    \caption{Performance comparisons with the DP-HOI VPF Operation in the HICO-DET DT mode.
    }
    \label{table:VPF_ablation_results}
    \centering
    \small
     \resizebox{0.35\textwidth}{!}{
      \begin{tabular}{c|ccc}
    \toprule
    Methods & Full & Rare & Non-Rare\\
    \midrule 
    w/o fusion &  33.07 &  28.96 &  34.30 \\
    avg-pooling & 32.76 & 28.40 & 34.06\\
    max-pooling & \textbf{33.34} & \textbf{29.89} & \textbf{34.37} \\
\bottomrule
    \end{tabular}
     }
     % \vspace{-4mm}
\end{table}

Second, ``avg-pooling'' means that we average all the RPQ prediction results and imposed supervision. Table~\ref{table:VPF_ablation_results} shows that this setting in the full HOI categories decreases by 0.58\% mAP. This may be because this setting implicitly forces all RPQs to make confident predictions according to the annotations.

\begin{table}[htbp]
    \caption{Performance comparisons with different parameter initializations. \textsuperscript{\dag} means DN was adopted during the fine-tuning stage.
    }
    \label{table:more_ initialization_results}
    \centering
    \small
     \resizebox{0.45\textwidth}{!}{
      \begin{tabular}{c|ccc}
        \toprule
        Methods & w/o decoder & w/o interaction decoder & full \\
        \midrule
        CDN-S\textsuperscript{\dag} & 31.06 & 31.98 & 31.39 \\
    CDN-S\textsuperscript{\dag} + Ours & \textbf{33.10} & \textbf{34.16} &\textbf{35.00}\\

        \bottomrule 
    \end{tabular}
     }
\end{table}

\noindent\textbf{The Comparisons with Various Model Initializations.}
As illustrated in Table~\ref{table:more_ initialization_results}, we compared the performance of three model initialization strategies using the CDN-S\textsuperscript{\dag} model: (a) without the decoder, initializing the backbone and encoder; (b) without the interaction decoder, initializing the backbone, encoder and detection decoder; (c) full, initializing the backbone, encoder, detection and interaction decoder.
CDN-S\textsuperscript{\dag} adopts the model pre-trained using the MS-COCO dataset. We also utilized the pre-trained detection decoder to initialize the CDN-S\textsuperscript{\dag} detection and interaction decoders in the (c) setting.   

As illustrated in Table~\ref{table:more_ initialization_results}, DP-HOI outperforms the baseline by 2.04\%, 2.18\% and 3.61\% mAP in the (a), (b) and (c) settings, respectively.
Moreover, when we only initialized our pre-trained model’s backbone and encoder, the performance reached 33.10\% mAP, outperforming the original pre-trained model with any of the three initialization strategies.
These results demonstrate that DP-HOI incorporates action-related features in the backbone and encoder, enhancing our pre-trained model's applicability across various HOI models.
\vspace{3mm}

\begin{figure}[t]
  \begin{center}
  \includegraphics[width=0.95\linewidth]{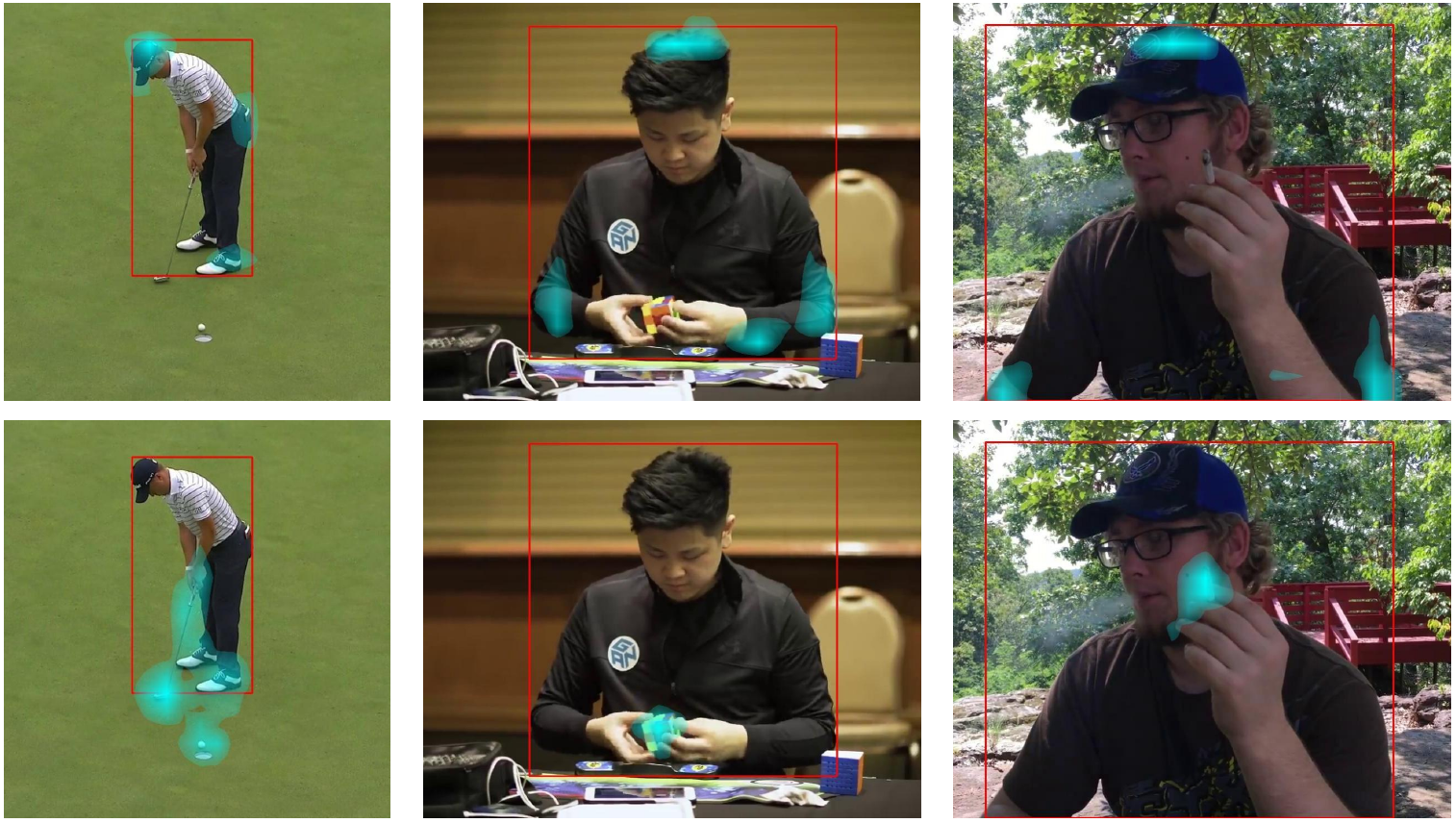}
  \end{center}
  \caption{ 
  Visualization of the attention maps in the decoder layers.
  The two rows represent results for the detection and interaction decoders, respectively.
  }
  \vspace{-3mm}
  \label{fig:visual-dphoi}
\end{figure}

% \vspace{-3mm}
\subsection{The DP-HOI Visualizations}\label{sec:Visualizations}
% \vspace{-2mm}
As illustrated in Figure \ref{fig:visual-dphoi}, we visualized the attention maps for the last detection (i.e., the first row) and interaction decoder layers (i.e., the second row) of the most confident human query according to the RPQ. We observed that the detection attention maps accurately localize the person's boundaries. Likewise, the interaction attention maps accurately localize the interaction regions. Therefore, with the disentangled supervision signals, the two decoders use different features for object detection and interaction classification.

\section{Conclusions and Limitations}
In this paper, we addressed the pre-training problem for DETR-based HOI detection models.
Specifically, we proposed a disentangled pre-training framework that effectively explores readily available and large-scale object detection, action recognition and image-caption datasets. Our pre-training architecture is consistent with the downstream HOI detection task, facilitating efficient knowledge transfer. 
In addition, we conducted comprehensive experiments on two popular HOI detection benchmarks. 
The experimental results demonstrated our methods' superiority. A possible limitation of this study is that it requires GPUs with relatively large memories for pre-training. 
In the future, we will explore more efficient pre-training strategies, that can include more object detection, action recognition and image-caption datasets, further enhancing the pre-training stage.

\noindent\textbf{Broader Impacts.}
DP-HOI significantly improves the performance of HOI detection models. It could impact human-centric vision applications such as driver monitoring and health care systems. To the best of our knowledge, this study does not have any obvious negative social impacts.

\noindent\textbf{Acknowledgement.}
We thank Ziliang Chen for insightful discussions. 
This work was supported by the National Natural Science Foundation of China under Grant 62076101, Guangdong Basic and Applied Basic Research Foundation under Grant 2023A1515010007, the Guangdong Provincial Key Laboratory of Human Digital Twin under Grant 2022B1212010004, CAAI-Huawei MindSpore Open Fund and TCL Young Scholars Program. 

\clearpage
\setcounter{page}{1}
\maketitlesupplementary
\appendix

This supplementary material includes four sections. 
Section~\ref{sec:datasets details} provides more details about pre-training datasets. 
Section~\ref{sec:training details} describes more training details of DP-HOI.
We provide more experimental results on HOI detection methods in Section~\ref{sec:more hoi application}.
Section \ref{sec:zero-shot hoi} provides experimental results on various zero-shot settings, i.e., UV, RF-UC, and NF-UC.  

\vspace{1mm}
\section{More Details of Pre-training Datasets}
\vspace{1mm}
\label{sec:datasets details}
\noindent\textbf{Objects365~\cite{shao2019objects365}.} Objects365 is a large-scale object detection dataset, which contains nearly 1,724K images with annotations for object detection only.
From the 365 classes in Objects365, we select the classes that are overlapped with the 80 classes in COCO. Subsequently, we randomly sampled 117,266 images in the selected object classes.

\noindent\textbf{Haa500~\cite{chung2021Haa500}.} Haa500 is a video-based action recognition dataset. For each long video, we conduct sampling uniformly with a time interval of 0.5s. For each video that is shorter than 2s, we uniformly sample 4 frames in the video. 
Since the action changes in each video are very small, we utilized the sampled 526,44 video frames as an image-based action recognition database for pre-training.

\noindent\textbf{Kinetics-700~\cite{Carreira2019short}.} Kinetics-700 is a large-scale video-based action recognition dataset, which contains over 650K videos in 700 classes. 
For each long video, we randomly select a starting frame and sample 16 frames with a frame interval of 4. We uniformly sample videos in these 700 classes and obtain 117K videos.

\begin{figure*}[htbp]
  \begin{center}
  \includegraphics[width=1.0\linewidth]{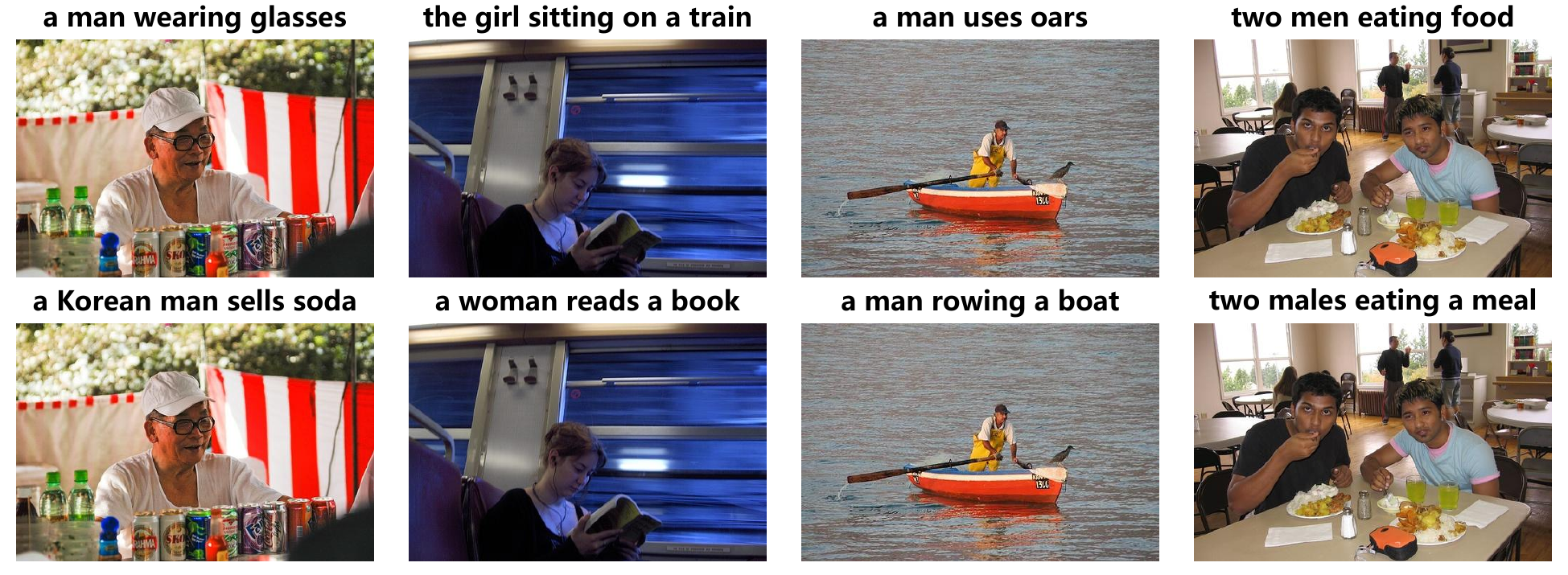}
  \end{center}
  % \vspace{-8mm}
  \caption{ 
  Visualization of obtained HOI triplets on Flickr30k. Each column indicates an image and its obtained HOI triplets in Flickr30k.  
  }
  \label{fig:data visualize}
\end{figure*}

\noindent\textbf{Flickr30k~\cite{young2014image}.} Flickr30k dataset contains nearly 30K images collected from Flickr. Each image owns 5 different captions. 
We use our rule-based language parser \cite{qi2018learning} to obtain qualified HOI triplets from captions. For example, given an image with captions \{``a man drives a car'', ``car runs on the road'', ``a man on the road''\}, we remove the triples where the subject is not a person and the relation is not a verb, i.e. \{``car runs on the road'', ``a man on the road''\}. We visualize some examples of obtained HOI triplets in Figure~\ref{fig:data visualize}. 
In the first three columns, the obtained HOI triplets exhibit diverse actions. As shown in the last column, there are several HOI triplets with similar semantics in our data. Different HOI triplets of similar meaning could enrich the diversity of text embeddings, which helps to increase the robustness of the model and prevent overfitting. Therefore, we do not perform additional processing for these synonyms.

\noindent\textbf{Visual Genome~\cite{krishna2017visual}.} Visual Genome(VG) consists of 101,174 images sampled from MS-COCO~\cite{lin2014microsoft}, with densely annotated object, attribute and relationship labels. 
We utilize the captions provided in VG and search for effective HOI triplets according to the same role for Flickr30k.
In addition, we do not utilize the VG images that are overlapped with the V-COCO test set to avoid information leakage.

\vspace{1mm}
\section{More Training Details}
\vspace{1mm}
\label{sec:training details}
We adopt the denoising (DN) strategy~\cite{pan2020spatio} to accelerate the pre-training and fine-tuning stages. In the pre-training stage, we first add noise to the ground-truth coordinates of each object bounding box and then use a two-layer FFN with ReLU to encode the coordinates~\cite{qu2022distillation}. We also used the label denoising strategy strategy in ~\cite{li2022dn} to speed up pre-training.
In the fine-tuning stage, we adopt the ground-truth coordinates of labeled human-object pairs to construct an auxiliary group of queries. Specifically, we add noise to the ground-truth coordinates of each human-object pair. We then adopt the encoding method proposed in~\cite{qu2022distillation} to obtain the auxiliary group of queries. 

The obtained auxiliary group of queries and the original group of learnable queries are fed into the decoder for prediction. This enables DETR-based models to converge more quickly~\cite{li2022dn,qu2022distillation}. For simplicity, the label denoising strategy in~\cite{li2022dn} is not used in fine-tuning stages. 

Moreover, data augmentation strategies are different for image datasets and video datasets.
As to image datasets, it includes random scaling, random horizontal flipping, random color jittering and gaussian blurring. The input images are resized to at least 800 pixels on the short size and at most 1333 pixels on the long side. 
As to video datasets, it includes random scaling, random cropping and random horizontal flipping. The spatial resolution of the input frames is set to 256 × 256.

Pre-training lasts for 200 epochs according to the MS-COCO dataset. The action datasets, including the action recognition and image caption datasets, are added in the 150th epoch. In each batch, the number of samples from object detection and action datasets is the same. When training with both action recognition and image-caption data, we keep the sampling ratio of object detection, action recognition and image-caption data as 2:1:1.

%%%%%%hico%%%%%%%%%%%%%%%%%%%%%%%%%
%%%%%%HICO%%%%%%%%%%%%%%%%%%%%%%%%%
\begin{table*}[htbp]
\begin{center}
\small
\caption{Performance comparisons on HICO-DET. GEN$_s$ denotes that distillation via CLIP is removed from GEN-VLKT$_s$. \textsuperscript{\dag} means DN is adopted in the fine-tuning stage. * denotes using a data augmentation strategy~\cite{qu2022distillation}. }
\resizebox{0.9\textwidth}{!}{
\begin{tabular}{c||c|ccc|ccc}
\toprule
\multirow{2}{*}{Methods}  & \multirow{2}{*}{Backbone}  & \multicolumn{3}{c|}{DT Mode} & \multicolumn{3}{c}{Known Object}\\
 && Full & Rare & Non-Rare & Full & Rare & Non-Rare  \\
\midrule
\midrule

UnionDet \cite{kim2020uniondet}& ResNet-50-FPN  &17.58  & 11.72  & 19.33 &19.76&14.68&21.27 \\
DRG \cite{gao2020drg}	&ResNet-50-FPN			&19.26	&17.74	&19.71	&23.40	&21.75	&23.89\\
PD-Net \cite{zhong2020polysemy}  & ResNet-152  &20.81 &15.90 &22.28 &24.78&18.88&26.54\\
PPDM \cite{liao2020ppdm}& Hourglass-104   & 21.73 &13.78 &24.10 &24.81&17.09&27.12\\
GGNet \cite{zhong2021glance}    & Hourglass-104   &23.47 &16.48 &25.60 &27.36&20.23&29.48\\

HOI-Trans \cite{zou2021end}   & ResNet-50  & 23.46 & 16.91 & 25.41 &26.15&19.24&28.22 \\
 AS-Net \cite{kim2020uniondet}   & ResNet-50  &28.87  & 24.25  & 30.25 &31.74&27.07&33.14\\
\midrule
QPIC  \cite{tamura2021qpic}  & ResNet-50   &29.07 &21.85 &31.23 &31.68&24.14&33.93 \\
QPIC+Ours & ResNet-50   &\textbf{30.63} &\textbf{25.27} &\textbf{32.23} &\textbf{32.94}&\textbf{27.24}&\textbf{34.64} \\
\midrule
CDN-S \cite{zhang2021mining}     & ResNet-50  &31.44 & 27.39 & 32.64 &34.09&29.63&35.42 \\
CDN-S\textsuperscript{\dag} \cite{zhang2021mining}    & ResNet-50  &31.98 & 28.61 & 32.99 &34.77&31.34&35.80 \\
CDN-S+Ours &ResNet-50  &34.27 &30.02	&35.54	&37.05	&33.09	&38.23\\
CDN-S\textsuperscript{\dag}+Ours &ResNet-50 &35.00 &32.38&\textbf{35.78}	&37.83&35.43	&\textbf{38.54}\\
CDN-S\textsuperscript{\dag}+CCS\textsuperscript{$*$}+Ours &ResNet-50  &\textbf{35.38} &\textbf{34.61}	&35.61	&\textbf{38.21}	&\textbf{37.43}	&38.44\\
\midrule
GEN$_s$+VLKT \cite{liao2022gen} & ResNet-50  &33.75&29.25&35.10&36.78&32.75&37.99 \\
GEN$_s$+Ours &ResNet-50  &\textbf{34.40} &\textbf{31.17}	&\textbf{35.36}	&\textbf{38.25}	&\textbf{35.64}	&\textbf{39.03}\\
\midrule
HOICLIP \cite{Ning_2023_CVPR} & ResNet-50  &34.69&31.12&35.74&37.61&34.47&38.54 \\
HOICLIP+Ours &ResNet-50  &\textbf{36.56} &\textbf{34.36}	&\textbf{37.22}	&\textbf{39.37}	&\textbf{36.59}	&\textbf{40.20}\\
\bottomrule
\end{tabular}}
\label{tab:hico_KO}
\end{center}
\end{table*}
%%%%%%%%%%%%%%%%%%%%%%%%%%%%%%%%%%%%%%%%%%%%

\vspace{1mm}
\section{More Experimental Results on HICO-DET}
\vspace{1mm}
\label{sec:more hoi application}
In this section, we demonstrate the effectiveness of DP-HOI in the Known-Object(KO) mode under default setting.

As shown in Table \ref{tab:hico_KO}, DP-HOI significantly boosts HOI detection performance on both DT and KO modes. 
When the pre-trained DETR weights by DP-HOI are applied to QPIC \cite{tamura2021qpic}, CDN-S\textsuperscript{\dag} \cite{zhang2021mining}, GEN$_s$+VLKT \cite{liao2022gen} and HOICLIP \cite{Ning_2023_CVPR}, we observe consistent performance gains by 1.26\, 3.06\%, 1.47\% and 1.76\% mAP in KO mode for the full categories. Moreover, the performance of QPIC, CDN-S\textsuperscript{\dag}, GEN$_s$+VLKT and HOICLIP on the rare HOI category is promoted by 3.42\%, 4.09\%, 2.89\% and 2.12\%, respectively.

%%%%%%%%%%%zero-shot%%%%%%%%%%% 
\begin{table*}[t]
    \caption{Application to zero-shot HOI detection on HICO-DET. GEN$_s$ denotes distillation via CLIP is removed from GEN-VLKT$_s$\cite{liao2022gen}.
    }
    \label{table:zero-shot gen}
    \centering
    \large
     \resizebox{0.9\textwidth}{!}{
      \begin{tabular}{c | ccc | ccc | ccc}
        \toprule
        \multirow{2}{*}{Methods}  & 
        \multicolumn{3}{c|}{UV}  & \multicolumn{3}{c|}{RF-UC} &
        \multicolumn{3}{c}{NF-UC} \\
        & Unseen & Seen & Full &
        Unseen & Seen & Full & Unseen & Seen & Full \\
        \midrule
        GEN$_s$+VLKT \cite{liao2022gen} 
                                         & 20.96 & 30.23 & 28.74 
                                         & 21.36 & 32.91 & 30.56
                                         & 25.05 & 23.38 & 23.71 \\
        GEN$_s$+Ours        
                            & \textbf{23.01} & \textbf{31.29} & \textbf{30.13}
                            & \textbf{23.73} & \textbf{33.59} & \textbf{31.61}
                            & \textbf{25.78} & \textbf{25.05} & \textbf{25.20} \\
        Improvement             
                            & +2.05 & +1.06 & +1.39
                            & +2.37 & +0.68 & +1.05
                            & +0.73 & +1.67 & +1.49 \\
        \midrule
        HOICLIP  \cite{Ning_2023_CVPR}   
                                         & 24.30 & 32.19 & 31.09
                                         & 25.53 & 34.85 & 32.99
                                         & 26.39 & 28.10 & 27.75 \\
        HOICLIP+Ours  
                      & \textbf{26.30} & \textbf{34.49} & \textbf{33.34}
                      & \textbf{30.49} & \textbf{36.17} & \textbf{35.03}
                      & \textbf{28.87} & \textbf{29.98} & \textbf{29.76}\\
        Improvement             
                            & +2.00 & +2.30 & +2.25
                            & +4.96 & +1.32 & +2.04
                            & +2.48 & +1.88 & +2.01 \\
        \bottomrule 
    \end{tabular}
     }
\end{table*}

% \vspace{-3mm}
\section{Zero-shot HOI Detection}
\label{sec:zero-shot hoi}
% \vspace{-2mm}
 In this section, we conduct experiments on three zero-shot settings, i.e. Unseen Verb (UV), Rare First Unseen Combination (RF-UC), and Non-rare First Unseen Combination (NF-UC), following previous work \cite{liao2022gen,Ning_2023_CVPR}. 

 We adopt GEN$_s$ \cite{liao2022gen} and HOICLIP \cite{Ning_2023_CVPR} as our baseline to verify the performance of DP-HOI on zero-shot settings.  For clean comparison, we follow the data split protocol on each zero-shot setting in the original papers \cite{liao2022gen,Ning_2023_CVPR}.

As illustrated in Table \ref{table:zero-shot gen}, our DP-HOI outperforms the baseline on most zero-shot settings. 
Compared with GEN$_s$+VLKT \cite{liao2022gen}, we achieve an impressive 2.05\%, 2.37\% and 0.73\% mAP gain for unseen categories under UV, RF-UC and NF-UC settings. 
Compared with HOICLIP \cite{Ning_2023_CVPR}, we observe consistent performance gains by 2.00\%, 4.96\% and 2.48\% mAP for unseen categories under UV, RF-UC and NF-UC settings. 
These experimental results further demonstrate the effectiveness of our pre-trained weights.

\end{document}